\documentclass[runningheads]{llncs}
\usepackage[T1]{fontenc}
\usepackage{graphicx,verbatim}
\usepackage{amsmath,amssymb}
\usepackage{booktabs}
\usepackage{multirow}
\usepackage{xcolor}
\usepackage{hyperref}
\usepackage{marvosym}
\usepackage{esvect}

\newcommand{\ours}{HiPath}
\newcommand{\cmark}{\checkmark}
\newcommand{\xmark}{$\times$}

\newcommand{\eg}{\textit{e.g.}}

\newcommand{\pp}{\text{pp}}

\begin{document}
\titlerunning{HiPath: Hierarchical VL Alignment for Pathology Diagnosis}
\title{\ours{}: Hierarchical Vision-Language Alignment for Structured Pathology Report Prediction}

\author{Ruicheng Yuan\inst{1,2} \and
Zhenxuan Zhang\inst{2} \and
Anbang Wang\inst{2} \and
Liwei Hu\inst{2} \and
Xiangqian Hua\inst{3} \and
Yaya Peng\inst{4} \and
Jiawei Luo\inst{1}\textsuperscript{(\Letter)} \and
Guang Yang\inst{2}\textsuperscript{(\Letter)}}
%index{Yuan, Ruicheng}
%index{Zhang, Zhenxuan}
%index{Wang, Anbang}
%index{Hu, Liwei}
%index{Hua, Xiangqian}
%index{Peng, Yaya}
%index{Luo, Jiawei}
%index{Yang, Guang}

%
\authorrunning{R. Yuan et al.}
% First names are abbreviated in the running head.
% If there are more than two authors, 'et al.' is used.
%
\institute{
College of Computer Science and Electronic Engineering, Hunan University, Changsha, Hunan, China\\
\email{\{raytion,luojiawei\}@hnu.edu.cn}
\and
Department of Bioengineering and Imperial-X, Imperial College London, London, UK\\
\email{\{r.yuan25,g.yang\}@imperial.ac.uk}
\and
Department of Pathology, Xiangtan Maternal and Child Health Hospital, Xiangtan, Hunan, China
\and
Department of Pathology, The First People’s Hospital of Xiangtan City, Xiangtan, Hunan, China
}

\maketitle

% ============================================================
% ABSTRACT
% ============================================================
\begin{abstract}
Pathology reports are structured, multi-granular documents encoding diagnostic conclusions, histological grades, and ancillary test results across one or more anatomical sites; yet existing pathology vision-language models (VLMs) reduce this output to a flat label or free-form text. We present \ours{}, a lightweight VLM framework built on frozen UNI2 and Qwen3 backbones that treats structured report prediction as its primary training objective. Three trainable modules totalling 15\,M parameters address complementary aspects of the problem: a Hierarchical Patch Aggregator (HiPA) for multi-image visual encoding, Hierarchical Contrastive Learning (HiCL) for cross-modal alignment via optimal transport, and Slot-based Masked Diagnosis Prediction (Slot-MDP) for structured diagnosis generation. Trained on 749\,K real-world Chinese pathology cases from three hospitals, \ours{} achieves 68.9\% strict and 74.7\% clinically acceptable accuracy with a 97.3\% safety rate, outperforming all baselines under the same frozen backbone. Cross-hospital evaluation confirms generalisation with only a 3.4\pp{} drop in strict accuracy while maintaining 97.1\% safety. Code and evaluation protocol:
\url{https://github.com/raytions/hipath}.
\end{abstract}

% ============================================================
% 1. INTRODUCTION
% ============================================================
\section{Introduction}
\label{sec:intro}

Recent progress in computational pathology has been driven by large-scale  self-supervised foundation models~\cite{chen2024uni,vorontsov2024virchow,karasikov2025midnight}  and vision-language models (VLMs)~\cite{huang2023plip,ikezogwo2024quilt}  that connect histopathology images with clinical text.  These models have become strong generic feature extractors and support a wide  range of downstream tasks, including tissue classification, biomarker  prediction, and zero-shot recognition~\cite{lu2024conch,huang2023plip}.

However, the clinical output of pathology is rarely a single label.  A routine report is a structured record that jointly specifies (i) a primary  diagnosis, (ii) a histological grade, and (iii) immunohistochemistry (IHC)  results, often repeated across multiple anatomical sites for the same  patient~\cite{who2019tumours}.  This structure is central to decision-making: different fields have different  clinical consequences, so an evaluation that collapses all errors into flat  accuracy can obscure safety-critical failures.  At the same time, open-ended text generation offers no guarantee that the  output is slot-consistent or reliably parseable.

This leaves an open question for pathology VLMs: can we train and evaluate a model whose primary objective is structured report prediction, rather than flat classification or unconstrained generation? To our knowledge, existing pathology VLMs are not trained or benchmarked for structured, slot-level diagnosis prediction in this setting. A related gap concerns language and reporting conventions. All major pathology VLMs to date are trained exclusively on English corpora~\cite{huang2023plip,lu2024conch,ikezogwo2024quilt}, despite substantial clinical demand in non-English contexts. China accounts for approximately 24\% of global cancer cases~\cite{bray2024global,han2024cancer} and faces an estimated shortage of 90{,}000 pathologists~\cite{xu2020pathology}. Chinese reports also follow distinct conventions, including site-prefixed segment identifiers, standardised grading terminology, and structured IHC formatting. While Chinese groups have introduced notable MIL methods~\cite{shao2021transmil}, pathology foundation models~\cite{wang2024chief}, and VLMs~\cite{sun2024pathasst}, none target structured Chinese diagnosis prediction at scale.

\begin{figure}[t]
\includegraphics[width=\textwidth]{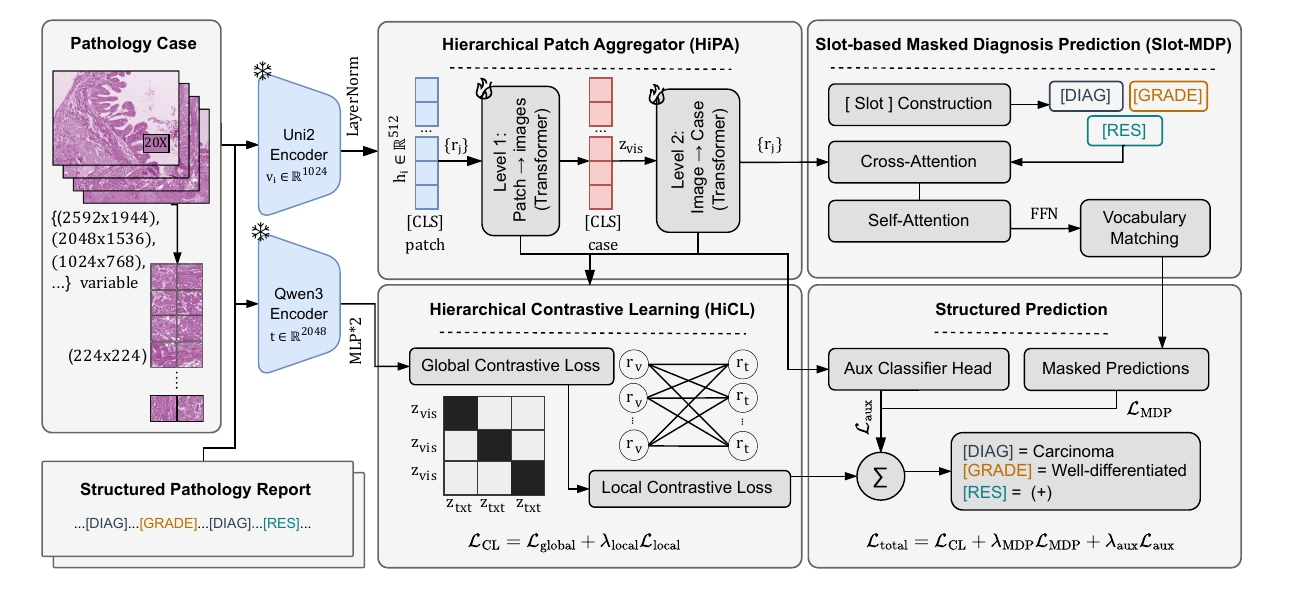}
\caption{Overview of \ours{}. Frozen UNI2 patch features and Qwen3
text embeddings are processed by three trainable modules: HiPA
aggregates patches hierarchically into case representations; HiCL
aligns modalities at case and segment level via optimal transport;
Slot-MDP fills typed diagnostic slots from visual features alone.
Free-text reports are used only during training.}
\label{fig:architecture}
\end{figure}

In this work, we propose \ours{} (Fig.~\ref{fig:architecture}), a lightweight framework that treats structured report prediction as the central learning objective. \ours{} is built on frozen UNI2~\cite{chen2024uni} and Qwen3~\cite{yang2025qwen3} backbones and adds 15\,M trainable parameters. The model (1) aggregates multi-image evidence selected by pathologists via hierarchical cross-attention (HiPA), (2) uses segment and case-level image-text alignment via optimal transport as an auxiliary training signal (HiCL)~\cite{cuturi2013sinkhorn}, and (3) predicts typed diagnostic slots by matching against a closed vocabulary in the frozen LLM embedding space (Slot-MDP). Free-text reports are used only during training; at inference, only frozen vocabulary embeddings and visual features are required. Trained on 749\,K cases from three hospitals, \ours{} achieves 68.9\% strict accuracy,  74.7\% clinically acceptable accuracy, and 97.3\% safety. Vision-only classifiers on the same frozen visual features plateau at 30--32\%, and a non-hierarchical variant with identical text access trails by 12.9\,\pp{}. Cross-hospital evaluation (training on two hospitals, testing on the third) shows a 3.4\,\pp{} drop while safety remains above 97\%. 

Our contributions are as follows.
(1)~We formalise structured slot-level report prediction as a primary objective for pathology VLMs and show that it substantially outperforms flat classification. (2)~We introduce a lightweight architecture on frozen backbones that combines hierarchical multi-image aggregation, OT-based local alignment, and vocabulary-grounded slot prediction for structured diagnosis. (3)~We propose a four-level clinical evaluation protocol with safety-aware metrics and validate it on 749\,K cases from three hospitals. The evaluation toolkit, including the 304-term vocabulary, synonym mappings, coarse taxonomy, and safety boundary definitions, is publicly released to facilitate reproducible research in structured pathology diagnosis.

% ============================================================
% 2. METHODS
% ============================================================
\section{Methods}
\label{sec:methods}

\textbf{Problem formulation.} Each case $c$ consists of $J$
physician-selected ROI images and a structured report parsed into
anatomical-site segments. In routine Chinese pathology practice,
reports follow a conventional structure organised by anatomical site,
with each site recording a primary diagnosis, an optional histological
grade, and any ancillary findings such as IHC results. We exploit this
regularity by parsing historical reports into segments and replacing
diagnostic terms with typed masks \texttt{[DIAG]}, \texttt{[GRADE]},
\texttt{[RES]}, yielding $K$ slots $\{(m_k,\tau_k)\}_{k=1}^{K}$,
where $m_k\in\mathcal{V}_{\tau_k}$ is the ground-truth term and
$\tau_k$ its type.

At inference, the model requires a report template specifying the
number of segments, slot positions, and slot types. This structure is
derivable from information available before pathologist examination:
the clinical requisition records the specimen source and number of
tissue blocks, while the IHC panel is ordered at accessioning. For
example, a breast biopsy with ER/PR testing yields a predictable
template of \texttt{[DIAG]}+\texttt{[GRADE]}+\texttt{[RES(ER)]}+%
\texttt{[RES(PR)]}, but this structure reveals
only which fields need to be filled, not whether the tissue is benign
or malignant, nor whether any marker is positive or negative.
Slot queries attend exclusively to visual features, so all diagnostic
predictions are grounded in morphological evidence.
In our experiments, templates are derived from the ground-truth report
structure; in deployment, they would be constructed from clinical
requisition and accessioning metadata.

\textbf{Training data.} We collected 749\,K pathology cases from three tertiary hospitals in China spanning 2011--2025.  Hospitals~A and C are general hospitals; Hospital~B is a maternal-child hospital. The top five organs are uterus (9.3\%), cervix (8.1\%), stomach (6.1\%), colon (5.0\%), and breast (3.1\%), covering 42 distinct organ sites. Each case pairs one or more pathologist-selected ROI images (median 1, P99${=}$8) with a structured Chinese report. The vocabulary was curated to 304 standardised terms (184~DIAG + 93~GRADE + 27~RES) after manual synonym consolidation by two senior pathologists and frequency filtering (${\geq}10$ occurrences); the top 10\% of terms cover 90.8\% of predictions. Cases were split 80/10/10 at the patient level independently per hospital, yielding approximately 600\,K/75\,K/75\,K cases for training, validation, and testing. The 304-term vocabulary and safety boundaries were defined based on clinical standards prior to splitting. This study was conducted under IRB approval; all data were de-identified.

% \begin{table}[t]
% \centering
% \caption{Dataset overview. Patient-level 80/10/10 split applied
% independently per hospital.}
% \label{tab:dataset}
% \begin{tabular}{lcccc}
% \toprule
%  & Hosp.\ A (General) & Hosp.\ B (Maternal) & Hosp.\ C (General)
%  & Total \\
% \midrule
% Cases        & ${\sim}$93\,K & ${\sim}$92\,K & ${\sim}$564\,K & 749\,K \\
% Organ sites  & 38 & 12 & 41 & 42 \\
% Multi-image  & \multicolumn{4}{c}{${\sim}$40\% of all cases} \\
% \bottomrule
% \end{tabular}
% \end{table}

\textbf{Feature extraction.} Visual features: frozen UNI2
(ViT-Giant/14) tiles each image into $224{\times}224$ patches and
encodes them to native 1536-d vectors, which we project to 1024-d. Text features: frozen Qwen3-30B-A3B~\cite{yang2025qwen3}
(MoE, 30B total/3B activated) encodes each report segment
independently, yielding 2048-d embeddings
$\{\mathbf{s}_l\}_{l=1}^{L}$ via mean pooling. Feature
pre-extraction cost ${\sim}$120~GPU-h (H100).

\subsection{Hierarchical Patch Aggregator (HiPA)}
\label{sec:hipa}

A pathology case often contains multiple ROI images capturing different
tissue regions. Unlike whole-slide methods that apply self-attention
within a single gigapixel image~\cite{chen2022hipt}, HiPA uses
cross-attention with learnable queries to aggregate across multiple
independent images selected by pathologists. Patch embeddings are
projected: $\mathbf{h}_i =
\mathrm{LN}(\mathbf{W}_\mathrm{proj}\mathbf{v}_i)\in\mathbb{R}^d$
($d{=}512$). Two cross-attention levels aggregate them:

\textbf{Level~1 (Patch$\to$Image).} A learnable \texttt{[CLS]} token
queries patches within image~$j$:
\begin{equation}
\mathbf{r}_j = \mathrm{TransBlock}([\mathbf{q}_\mathrm{patch};\,
\mathbf{h}_1,\ldots,\mathbf{h}_{N_j}])\big|_\texttt{CLS}
\end{equation}

\textbf{Level~2 (Image$\to$Case).} A second token aggregates
per-image representations:
\begin{equation}
\mathbf{z}_\mathrm{vis} =
\mathrm{TransBlock}([\mathbf{q}_\mathrm{case};\,
\mathbf{r}_1,\ldots,\mathbf{r}_J])\big|_\texttt{CLS}
\end{equation}
yielding case embedding $\mathbf{z}_\mathrm{vis}\in\mathbb{R}^d$ and
per-image representations $\{\mathbf{r}_j\}$ for downstream modules.

\subsection{Hierarchical Contrastive Learning (HiCL)}
\label{sec:hicl}

HiCL learns a shared vision-language embedding space via contrastive
alignment. It serves as an auxiliary training signal that regularises
HiPA's visual representations and enables image-text retrieval.
Qwen3 segment embeddings are projected via a two-layer MLP to
case-level $\mathbf{z}_\mathrm{txt}$ and segment-level
$\{\mathbf{s}_l\}_{l=1}^{L}$.

\textbf{Global loss.} Symmetric InfoNCE on case-level pairs:
\begin{equation}
\mathcal{L}_\mathrm{global} = \tfrac{1}{2}\bigl[
\mathrm{CE}(\mathbf{S}\cdot e^{\alpha},\mathbf{y}) +
\mathrm{CE}(\mathbf{S}^\top\cdot e^{\alpha},\mathbf{y})\bigr]
\end{equation}
where $S_{ij}=\bar{\mathbf{z}}_{\mathrm{vis},i}^\top
\bar{\mathbf{z}}_{\mathrm{txt},j}$ (L2-normalised), $\mathbf{y}$ is
the identity target, and $\alpha$ is a learnable log-temperature
clamped to $[1,2.7]$.

\textbf{Local loss.} Per-image $\{\mathbf{r}_j\}_{j=1}^{J_c}$ and
per-segment $\{\mathbf{s}_l\}_{l=1}^{L_c}$ embeddings form an
unordered many-to-many correspondence with $J_c \neq L_c$ in general.
We adopt optimal transport with entropic
regularisation~\cite{cuturi2013sinkhorn} rather than hard matching
(\eg, Hungarian algorithm, which requires $J_c = L_c$) or independent
dot-product attention~\cite{radford2021clip} (which treats each pair
independently and does not enforce a global assignment). Sinkhorn OT
produces a differentiable, dense transport plan that distributes
probability mass across all image-segment pairs while respecting
marginal constraints.
The cost matrix
$C_{jl}=1-\bar{\mathbf{r}}_j^\top\bar{\mathbf{s}}_l$ is solved with
Sinkhorn-Knopp ($\epsilon{=}0.05$,
3~iterations, uniform marginals). Let
$\mathcal{B}_\mathrm{multi}$ denote mini-batch cases with $J_c{>}1$
and $L_c{>}1$:
\begin{equation}
\mathcal{L}_\mathrm{local} =
\frac{1}{|\mathcal{B}_\mathrm{multi}|}
\sum_{c\in\mathcal{B}_\mathrm{multi}}
\frac{\langle\mathbf{T}^*_c,\mathbf{C}_c\rangle}{\min(J_c,L_c)}
\end{equation}
where $\mathbf{T}^*_c\in\mathbb{R}^{J_c \times L_c}$ is the optimal
transport plan. The combined loss is $\mathcal{L}_\mathrm{HiCL} =
\mathcal{L}_\mathrm{global} + 0.5\,\mathcal{L}_\mathrm{local}$.

\subsection{Slot-based Masked Diagnosis Prediction (Slot-MDP)}
\label{sec:slotmdp}

Slot-MDP is the primary prediction module. Unlike masked language
modelling over open vocabularies, it operates over a closed,
type-specific vocabulary: each slot predicts from
$\mathcal{V}_{\tau_k}$ via cosine matching against fixed Qwen3
embeddings $\mathbf{e}_w\in\mathbb{R}^{2048}$.

Each masked slot receives a learnable query $\mathbf{q}_k =
\mathbf{E}_\mathrm{pos}(k) + \mathbf{E}_\mathrm{type}(\tau_k)$.
Slots attend to per-image representations via cross-attention, then
communicate via self-attention to encourage diagnostic consistency:
\begin{align}
\hat{\mathbf{q}}_k &=
  \mathrm{CrossAttn}(\mathbf{q}_k,\{\mathbf{r}_j\}) + \mathbf{q}_k
  \\
\tilde{\mathbf{q}}_k &=
  \mathrm{SelfAttn}(\hat{\mathbf{q}}_k,\{\hat{\mathbf{q}}_{k'}\})
  + \hat{\mathbf{q}}_k
\end{align}
Each slot is projected to $\mathbf{p}_k =
\mathrm{FFN}(\tilde{\mathbf{q}}_k)$ and scored:
\begin{equation}
p(w\mid k) = \frac{\exp(\beta\cdot\bar{\mathbf{p}}_k^\top
\bar{\mathbf{e}}_w)} {\sum_{w'\in\mathcal{V}_{\tau_k}}
\exp(\beta\cdot\bar{\mathbf{p}}_k^\top
\bar{\mathbf{e}}_{w'})}
\end{equation}
where $\beta$ is learnable (initialised to 10). Training uses a focal
loss~\cite{lin2017focal} over the ground-truth term $m_k$ of each slot:
\begin{equation}
\label{eq:lmdp}
\mathcal{L}_\mathrm{MDP} = \frac{1}{K}\sum_{k=1}^{K}
\bigl(1 - p(m_k\mid k)\bigr)^{\gamma}\,
\mathrm{CE}_{\epsilon}\bigl(p(\cdot\mid k),\, m_k\bigr),
\end{equation}
with focusing parameter $\gamma{=}2$ and label-smoothed
cross-entropy $\mathrm{CE}_{\epsilon}$ ($\epsilon{=}0.1$).

\textbf{Training objective.} The total loss is
$\mathcal{L} = \mathcal{L}_\mathrm{HiCL}
+ 0.5\,\mathcal{L}_\mathrm{MDP}
+ 0.1\,\mathcal{L}_\mathrm{aux}$,
where $\mathcal{L}_\mathrm{aux}$ is BCE for coarse category prediction
(12~classes). Trainable parameters total 15.0\,M (HiPA~5.2\,M,
HiCL~4.1\,M, Slot-MDP~5.7\,M). We trained for 50 epochs with AdamW
(lr $10^{-4}$, cosine schedule, batch 32) on $4{\times}$H100
(${\sim}$5.6\,h). All experiments are repeated over three random seeds;
we report mean values (std ${\leq}0.3\pp$ for all metrics).

\textbf{Evaluation protocol.} We define four per-slot indicators in
a hierarchical cascade:
(1)~strict: exact match;
(2)~semantic: match after synonym mapping;
(3)~coarse: same category (12~classes);
(4)~safe: no crossing of clinically dangerous boundaries
(benign$\leftrightarrow$malignant, non-adjacent grade jump, IHC
polarity reversal).
These satisfy $\text{strict} \Rightarrow \text{semantic} \Rightarrow
\text{coarse}$ and $\text{strict} \Rightarrow \text{safe}$, but
coarse and safe are independent. Clinically acceptable accuracy is
defined as:
\begin{equation}
\label{eq:accept}
\text{Acc}_\text{accept} = \frac{1}{N}\bigl|\{k : \text{strict}(k)
\lor \text{semantic}(k)
\lor (\text{coarse}(k) \land \text{safe}(k))\}\bigr|
\end{equation}
An exact or synonym match is always acceptable; a coarse match is
acceptable only if safe. This reflects clinical
practice~\cite{who2019tumours,raab2005clinical}: identifying the
broad diagnostic category suffices for the correct clinical pathway,
whereas crossing the benign-malignant boundary causes direct harm.
Safety boundaries were defined by consensus of two senior pathologists
(${>}$15\,years each). All metrics are computed per slot
($N{=}109{,}523$ in the test set).

% ============================================================
% 3. RESULTS AND DISCUSSION
% ============================================================
\section{Results and Discussion}
\label{sec:results}

\begin{table}[t]
\centering
\caption{Main results (\%). Top-1 = strict exact match.
Accept.\ = clinically acceptable accuracy (Eq.~\ref{eq:accept}).
TP = trainable parameters. All UNI2-based methods share frozen
features, data, and split. $\dagger$~Uses Qwen3 text during training
only; all methods use only visual features and frozen vocabulary
embeddings at inference.
\ours{} Cross = trained on Hospitals~A+C, tested on Hospital~B.}
\label{tab:main}
\setlength{\tabcolsep}{3pt}
\begin{tabular}{lllcccc}
\toprule
Method & Encoder & TP & Top-1 & Top-5 & Accept. & Safety \\
\midrule
Text-only (no images) & Qwen3 & 0 & 21.52 & 54.27 & --- & --- \\
\midrule
\multirow{3}{*}{Linear probe}
  & PLIP  & 156\,K & 10.64 & 45.13 & 33.91 & 78.15 \\
  & CONCH & 156\,K & 14.63 & 55.71 & 42.68 & 85.62 \\
  & UNI2  & 312\,K & 30.64 & 71.81 & 58.54 & 87.50 \\
\midrule
ABMIL~\cite{ilse2018abmil} & UNI2 & 603\,K  & 32.37 & 73.54 & 60.29 & 87.21 \\
Deep MLP       & UNI2 & 15.9\,M & 32.49 & 73.86 & 61.76 & 87.46 \\
Flat CrossAttn$^\dagger$ & UNI2 & 15.7\,M & 55.98 & 83.83 & 68.92 & 92.18 \\
\midrule
\textbf{\ours{}}$^\dagger$ & UNI2 & 15.0\,M &
  \textbf{68.89} & \textbf{88.68} & \textbf{74.74} & \textbf{97.32} \\
\ours{} Cross$^\dagger$  & UNI2 & 15.0\,M &
  65.47 & 80.99 & 68.52 & 97.06 \\
\bottomrule
\end{tabular}
\end{table}

\begin{table}[t]
\centering
\caption{Per-slot-type performance breakdown for \ours{} (\%).
$N$ = number of test slots of each type. Vocab = vocabulary size.}
\label{tab:perslot}
\setlength{\tabcolsep}{4pt}
\begin{tabular}{llrcccc}
\toprule
Slot type & Vocab & $N$ & Strict & Semantic & Accept. & Safety \\
\midrule
DIAG  & 184 & 95{,}372 & 69.68 & 74.25 & 76.14 & 96.93 \\
GRADE & 93  & 4{,}600 & 71.30 & 76.11 & 76.11 & 100.00 \\
RES   & 27  & 9{,}551 & 59.88 & 59.98 & 59.98 & 100.00 \\
\midrule
All   & 304 & 109{,}523 & 68.89 & 73.08 & 74.74 & 97.32 \\
\bottomrule
\end{tabular}
\end{table}

\begin{table}[t]
\centering
\caption{Ablation study with component configuration.
\cmark/\xmark{} = present/removed. Flat CrossAttn from
Table~\ref{tab:main} included for reference. Removing HiPA also
disables local OT (no per-image representations to align).}
\label{tab:ablation}
\setlength{\tabcolsep}{2.5pt}
\begin{tabular}{lccccccccc}
\toprule
  & \multicolumn{2}{c}{HiPA} & \multicolumn{2}{c}{HiCL} &
  \multirow{2}{*}{\shortstack{Slot-\\MDP}} &
  \multirow{2}{*}{Top-1} & \multirow{2}{*}{Accept.} &
  \multirow{2}{*}{\shortstack{i2t\\R@1}} &
  \multirow{2}{*}{\shortstack{4+img\\Top-1}} \\
\cmidrule(lr){2-3} \cmidrule(lr){4-5}
  & L1 & L2 & Glob. & Local &  &  &  &  & \\
\midrule
\textbf{Full \ours{}}
  & \cmark & \cmark & \cmark & \cmark & \cmark
  & \textbf{68.9} & \textbf{74.7} & \textbf{21.2} & \textbf{55.1} \\
w/o HiCL
  & \cmark & \cmark & \xmark & \xmark & \cmark
  & 67.6\,{\scriptsize($-$1.3)} & 64.3\,{\scriptsize($-$10.4)}
  & 0.0 & 53.2\,{\scriptsize($-$1.9)} \\
w/o HiPA
  & \xmark & \xmark & \cmark & \xmark & \cmark
  & 67.5\,{\scriptsize($-$1.4)} & 63.3\,{\scriptsize($-$11.4)}
  & 22.6 & 51.4\,{\scriptsize($-$3.6)} \\
w/o Slot-MDP
  & \cmark & \cmark & \cmark & \cmark & \xmark
  & --- & 64.6\,{\scriptsize($-$10.1)} & 21.1 & --- \\
\midrule
Flat CrossAttn$^\dagger$
  & \xmark & \xmark & \cmark & \xmark & \cmark
  & 56.0 & 68.9 & 14.3 & 40.7 \\
\bottomrule
\end{tabular}
\end{table}

\textbf{Baseline design.} The baselines in Table~\ref{tab:main} are
designed to progressively isolate the factors contributing to
\ours{}'s performance. (i)~Linear probes (PLIP, CONCH, UNI2) and
ABMIL~\cite{ilse2018abmil} are vision-only flat classifiers over the
full 304-term vocabulary, establishing the ceiling achievable without
text supervision or structured prediction. Deep MLP, the
highest-capacity vision-only baseline, applies a 4-block 2048-d
LN/GELU/Dropout MLP with a 304-way focal cross-entropy head to
mean-pooled UNI2 features; the Text-only row instead classifies from
frozen Qwen3 report embeddings without any image input, probing the
language prior available in report text alone. English-pretrained encoders
(PLIP, CONCH) reach ${\leq}$14.6\% on Chinese data, reflecting the
combined effect of domain shift and label-taxonomy mismatch. UNI2 classifiers plateau at 30--32\% regardless of
capacity (312\,K $\to$ 15.9\,M parameters yields $+$1.9\pp),
indicating that frozen features alone are insufficient for structured
prediction. (ii)~Flat CrossAttn$^\dagger$ adds text-supervised training
and slot-based prediction (Slot-MDP + global contrastive loss) while
replacing the two-level HiPA with a single \texttt{[CLS]} over all
patches and removing local OT. It reaches 56.0\%, isolating the
combined contribution of text alignment and the slot formulation as
${\sim}$24\pp{} over the vision-only baselines. (iii)~The w/o~HiCL
variant in Table~\ref{tab:ablation} retains HiPA and Slot-MDP but
removes all contrastive text alignment, serving as a purely
vision-based structured predictor. It achieves 67.6\%, confirming
that the slot formulation with hierarchical aggregation is the primary
driver of performance. Both \ours{} and Flat CrossAttn use Qwen3 text
only during training (marked $\dagger$); at inference, only frozen
vocabulary embeddings and visual features are used. The remaining 12.9\pp{} gap between Flat CrossAttn and the
full model is decomposed by the ablation: hierarchical aggregation
(HiPA: $+$1.4\pp{} overall, $+$3.6\pp{} on 4+~image cases) and
contrastive alignment (HiCL: $+$1.3\pp{} strict, $+$10.4\pp{} acceptable).

\textbf{Per-slot-type analysis.} Table~\ref{tab:perslot} reports
performance broken down by slot type. DIAG slots, which draw from the
largest vocabulary (184 terms) and require the most nuanced visual
interpretation, show the lowest strict accuracy. GRADE and RES slots
both maintain 100\% safety: all grading errors are between adjacent
grades, and non-strict RES errors are predominantly semi-quantitative
confusions (\eg, (1+) vs.\ (2+)) that do not alter clinical
decisions. Among the 3.1\% of DIAG slots classified as unsafe, the top
confusions are: chronic inflammation$\leftrightarrow$dysplasia
(847~slots), adenoma$\leftrightarrow$adenocarcinoma (312~slots), and
leiomyoma$\leftrightarrow$leiomyosarcoma (201~slots). These are
morphologically challenging distinctions with documented high
inter-observer variability among expert
pathologists~\cite{raab2005clinical}.

\textbf{Cross-hospital generalisation.} \ours{} Cross is trained on
Hospitals~A+C and tested on Hospital~B, with no overlapping patients,
staining protocols, or scanners. It achieves 65.5\% strict accuracy
and 97.1\% safety (Table~\ref{tab:main}), only 3.4\pp{} below the
in-distribution result. Hospital~B is a maternal-child hospital with a
narrower organ distribution (12 sites vs.\ 38--41 for A and C), so
this experiment primarily tests generalisation to a different clinical
population and staining protocol rather than to entirely unseen organ
types.

\textbf{Long-tail analysis.} Head terms (top 10\%, covering 90.8\%
of predictions) achieve 70.2\% strict accuracy; mid-frequency terms
(109~terms) reach 44.4\%; tail terms (136~terms, 0.2\% of
predictions) drop to 2.6\%. Three rebalancing strategies
(square-root resampling, Focal loss adjustment, class-balanced
loss~\cite{cui2019class}) all degraded head accuracy by 2--5\pp{}
without meaningful tail improvement ($+$0.8\pp). Inspection of the
top-10 tail-term confusions confirms that they are predominantly
confused with morphologically similar head terms (\eg, rare carcinoma
subtypes misclassified as the dominant ``adenocarcinoma'' category),
consistent with the hypothesis that tail errors reflect visual
ambiguity rather than sampling imbalance.

\textbf{Ablation.} Table~\ref{tab:ablation} confirms that each module
targets a distinct capability. Removing HiCL collapses retrieval
(R@1: $21.2 \to 0$) and reduces acceptable accuracy by 10.4\pp{},
while strict accuracy drops more modestly ($-$1.3\pp) as Slot-MDP
compensates; HiCL's value is therefore primarily in cross-modal
alignment and the auxiliary regularisation it provides to visual
representations. Removing HiPA costs 3.6\pp{} on multi-image cases
(4+~images) but only 0.8\pp{} on single-image cases, confirming that
hierarchical aggregation specifically benefits multi-ROI
interpretation. The gap between Flat CrossAttn (56.0\%) and w/o~HiPA
(67.5\%) shows that HiPA and local OT interact: without per-image
representations, OT has nothing to operate on, creating a compounding
deficit.

We note two limitations of the current experimental design.
First, we do not compare against a generative baseline (\eg, a
visual-adapter-augmented Qwen3 with constrained decoding over the same
vocabulary). Such a comparison would require fine-tuning the LLM
backbone, which substantially exceeds the 15\,M parameter budget of
our framework; we consider this an important direction for future work.
Second, while we motivate Sinkhorn OT by its ability to handle
unequal-sized sets differentiably (\S\ref{sec:hicl}), a systematic
comparison against alternative soft-assignment strategies (\eg, softmax
attention) is not included.

% ============================================================
% 4. CONCLUSION
% ============================================================
\section{Conclusion}

We presented \ours{}, a lightweight vision-language framework that
treats structured pathology report prediction as its primary learning
objective. With only 15\,M trainable parameters on frozen backbones,
\ours{} achieves 68.9\% strict accuracy, 74.7\% clinically acceptable
accuracy, and 97.3\% safety on 749\,K cases, with robust
cross-hospital generalisation. Several limitations point to future
work: the current evaluation covers three hospitals in one region
(broader multi-centre validation is needed); the model operates on
pathologist-selected ROIs rather than whole-slide images; and cases
outside the 304-term vocabulary are assigned the nearest term rather
than rejected. We release the evaluation protocol, including the
vocabulary, synonym mappings, coarse taxonomy, and safety boundary
definitions, to support reproducibility and further research.

% ============================================================
% ACKNOWLEDGMENTS
% ============================================================
% NOTE: \begin{credits} with \ackname/\discintname requires the current
% Springer llncs.cls (2022+), used by the official MICCAI camera-ready
% template. If an older llncs.cls is used, replace the block below with:
%   \subsubsection*{Acknowledgments.} ...funding text...
%   \subsubsection*{Disclosure of Interests.} ...statement...
\begin{credits}
\subsubsection{\ackname}
R.~Yuan was supported by the China Scholarship Council 
(CSC). G.~Yang was supported in part by
the ERC IMI (101005122), the H2020 (952172), the MRC (MC/PC/21013),
the Royal Society (IEC/NSFC/211235), the NVIDIA Academic Hardware
Grant Program, the SABER project supported by Boehringer Ingelheim
Ltd, the NIHR Imperial Biomedical Research Centre (RDA01), the
Wellcome Leap Dynamic Resilience program (co-funded by Temasek Trust),
UKRI guarantee funding for Horizon Europe MSCA Postdoctoral
Fellowships (EP/Z002206/1), a UKRI MRC Research Grant, TFS Research
Grants (MR/U506710/1), the Swiss National Science Foundation (Grant
No.~220785), and the UKRI Future Leaders Fellowship (MR/V023799/1,
UKRI2738). J.~Luo was supported in part by the National Natural Science
Foundation of China (No.~62372165).

\subsubsection{\discintname}
The authors have no competing interests to declare that are relevant
to the content of this article.
\end{credits}

% ============================================================
% REFERENCES
% ============================================================
\bibliographystyle{splncs04}
\bibliography{reference}

\end{document}